\documentclass[letterpaper, 10 pt, conference]{ieeeconf}  

\IEEEoverridecommandlockouts                              

\title{\LARGE \bf
Closing Sim-to-Real Gap for Heavy-loaded Humanoid Agile Motion Skills via Differentiable Simulation
}

\author{Xingyi Wang$^{1,2*}$, Chenyun Zhang$^{2*}$, Weiji Xie$^{2,3*}$, Chao Yu$^{4}$, Wei Song$^{1}$, Chenjia Bai$^{2\dagger}$, Shiqiang Zhu$^{1\dagger}$
\vspace{0.5em}
\\
$^{1}$Zhejiang University
\quad
$^{2}$Institute of Artificial Intelligence (TeleAI), China Telecom
\\
$^{3}$Shanghai Jiao Tong University
\quad
$^{4}$Lumos Robotics
\vspace{0.5em}
\\
\small{*Equal contribution \quad $\dagger$Corresponding author}
}

\usepackage{cite}
\usepackage{xcolor}
\usepackage{amssymb}
\usepackage{booktabs}    
\usepackage{amsmath}     
\usepackage{siunitx}     
\usepackage{tabularx} 
\usepackage{graphicx}
\usepackage{stfloats}
\usepackage{ragged2e}
\usepackage[table]{xcolor}
\usepackage{makecell}

\usepackage{caption}

\makeatletter
\let\NAT@parse\undefined
\makeatother

\usepackage{hyperref}

\hypersetup{
    colorlinks=true,        
    linkcolor=blue,         
    citecolor=green,        
    urlcolor=magenta,       
    pdfborder={0 0 0}       
}
\newcommand{\authnote}[2][]{%
  \ifx&#1&%
    $\ll$\textsf{\footnotesize #2}$\gg$
  \else
    $\ll$\textsf{\footnotesize #1$\Vert$#2}$\gg$
  \fi
}

\begin{document}

\maketitle
\thispagestyle{empty}
\pagestyle{empty}



\begin{abstract}

Humanoid robots deployed in real-world scenarios often need to carry unknown payloads, which introduce significant mismatch and degrade the effectiveness of simulation-to-reality reinforcement learning methods. To address this challenge, we propose a two-stage gradient-based system identification framework built on the differentiable simulator MuJoCo XLA. The first stage calibrates the nominal robot model using real-world data to reduce intrinsic sim-to-real discrepancies, while the second stage further identifies the mass distribution of the unknown payload. By explicitly reducing structured model bias prior to policy training, our approach enables zero-shot transfer of reinforcement learning policies to hardware under heavy-load conditions. Extensive simulation and real-world experiments demonstrate more precise parameter identification, improved motion tracking accuracy, and substantially enhanced agility and robustness compared to existing baselines. \href{https://mwondering.github.io/halo-humanoid/}{Project Page: HALO-Humanoid}
\end{abstract}

\section{INTRODUCTION}

Humanoid robots are increasingly expected to move beyond controlled laboratory settings and perform complex physical tasks in unstructured environments, such as industrial facilities and disaster-response sites. In these scenarios, robots are typically required to manipulate tools, carry equipment, or integrate additional functional modules to accomplish task-specific objectives. Such loads induce substantial shifts in the robot’s system dynamics by altering its total mass, Center of Mass (CoM), and inertia distribution \cite{lee2020adaptive}. Although recent advances in deep reinforcement learning (RL) have achieved impressive agility \cite{liao2025beyondmimic,experiment-datasets-hub,xie2025kungfubot,han2025kungfubot2}, policies trained with nominal dynamics typically degrade when exposed to large, structured payload variations.

We interpret heavy-load carrying as a structured sim-to-real gap problem. Unlike small stochastic noise, payload changes introduce systematic and large-scale dynamic mismatch between the simulation and the physical robot. A widely adopted strategy to mitigate the sim-to-real gap is Domain Randomization (DR) \cite{related-work-DR1, related-work-DR2}, which improves robustness by exposing policies to diverse, randomized dynamics during training. However, by treating payload variations as disturbances to hedge against, DR often yields conservative behaviors and cannot fully exploit the robot’s physical capabilities.

Another line of work adopts data-driven approaches that leverage real-world interaction data to reduce model mismatch. Within this category, System Identification (SysID) plays a central role, which directly estimates physical parameters—such as mass, CoM location, and inertia—using real interaction data. However, traditional SysID faces several practical limitations for floating-based systems like humanoid robots. They often rely on specialized sensing equipment, such as joint torque sensors~\cite{hwangbo2019learning} or external motion capture systems~\cite{related-work-sampling-based-2,sontakke2023residual}, which limits applicability. Moreover, many procedures require isolating subsystems or partially disassembling the robot~\cite{intro-ETH,kovalev2025achieving,tan2018sim}, making them impractical for complex platforms.

\begin{figure}[t] 
    \centering
    \includegraphics[width=\linewidth]{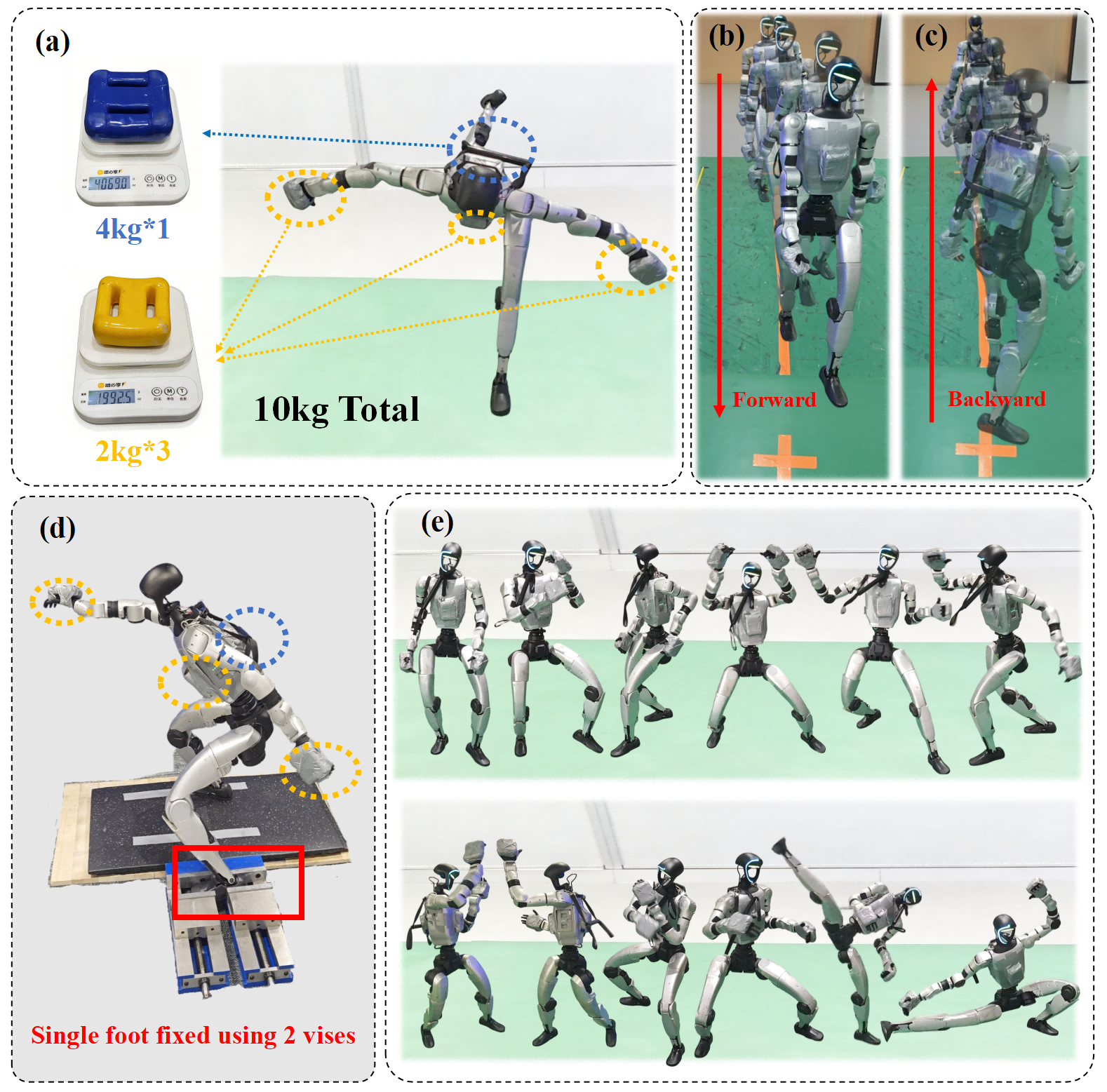}
    \caption{\textbf{Performance of HALO in heavy-loaded scenarios.}
    \textbf{(a)} The payload settings used by HALO.
    \textbf{(b, c)} Compared to DR, HALO significantly improves the accuracy of straight-line bidirectional walking.
    \textbf{(d)} Data collection under payload conditions with a single-foot constraint.
    \textbf{(e)} HALO enables challenging humanoid locomotion skills.}
    \label{fig:teaser}
\end{figure}

\begin{figure*}[t]
    \centering
    \includegraphics[width=0.95\textwidth]{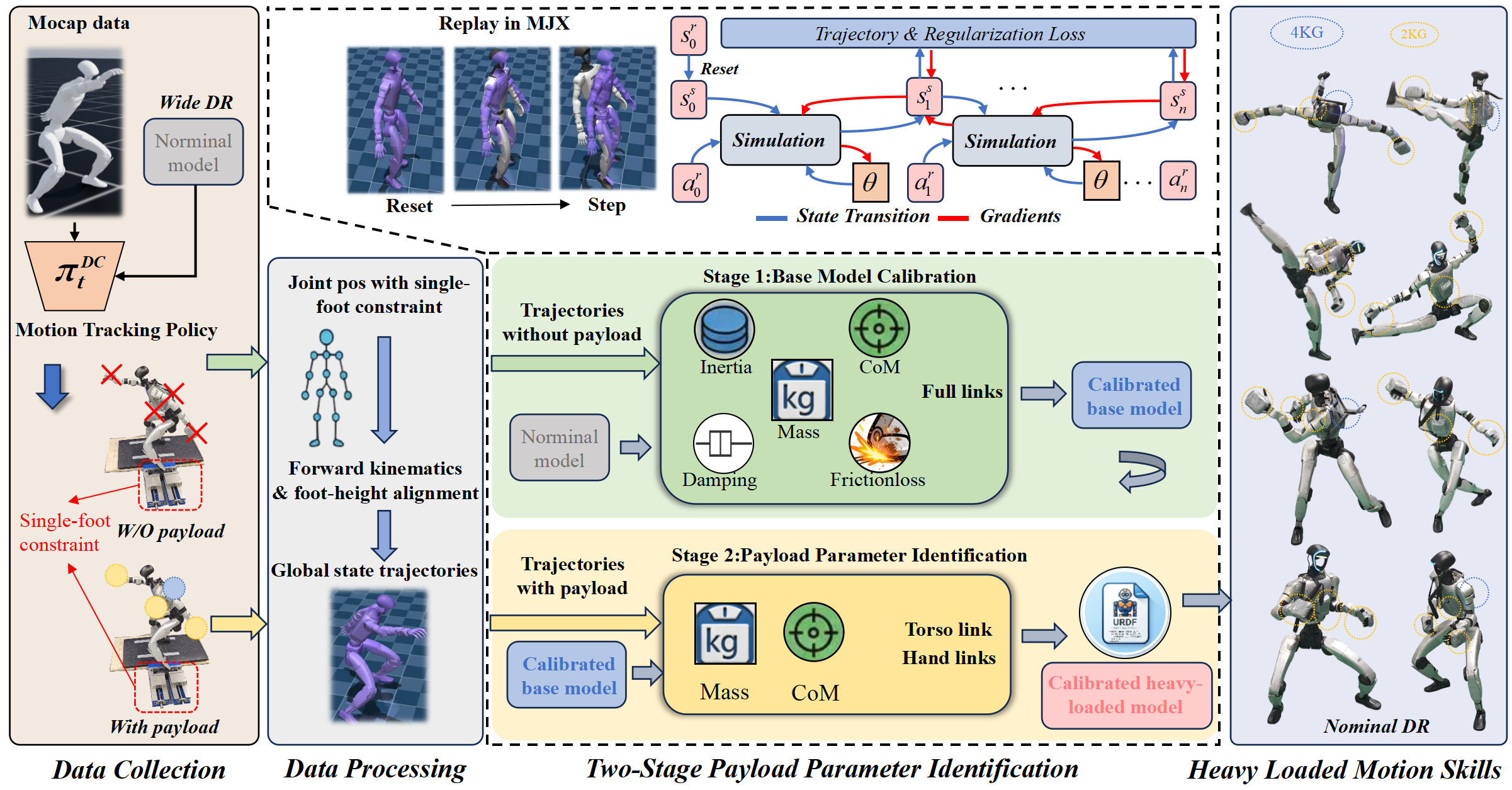} 
    \caption{\textbf{Overview of HALO.} 
    \textbf{(a) \textit{Data Collection:}} Trajectories are collected under both loaded and unloaded conditions using exploration policy trained with wide DR, followed by real-world deployment with a fixed foot constraint.
    \textbf{(b) \textit{Data Processing:}} Full-body trajectories reconstruction from joint-state measurements via forward kinematics and foot-height alignment. 
    \textbf{(c) \textit{Two-stage Payload-related Parameter Identification:}} Stage 1 optimize the full set of model parameters to yield a calibrated base model using trajectories without payload. Based on the calibrated model, stage 2 optimize only the payload-related parameters, using trajectories collected under loaded conditions.
    \textbf{(d) \textit{Heavy-loaded Motion Skills:}} The accurate identified model parameters enabling zero-shot sim-to-real transfer of the learned skills to the physical heavy-loaded humanoid.}
    \label{fig:method_overview}
\end{figure*}

To overcome these limitations, we propose \textbf{HALO} (\textbf{H}e\textbf{A}vy-\textbf{LO}aded humanoid motion control), a unified framework that seamlessly integrates differentiable simulation based SysID into the reinforcement learning control pipeline for heavy-loaded humanoids. HALO formulates parameter identification as a trajectory-level optimization problem inside a differentiable physics simulator MuJoCo XLA, leveraging the analytical gradients over multiple simulation steps to enable efficient gradient-based optimization of physical parameters directly from real-world interaction data.

To make such identification practical on floating-base humanoids without relying on external sensing systems, we introduce a lightweight data collection procedure in which one foot is mechanically constrained during trajectory acquisition. This constraint removes the ambiguity in global base drift and enables reliable reconstruction of full-body motion from onboard joint encoders.

We also identify a central challenge in payload identification: the entanglement between nominal model mismatch and load-induced model change. Without separating these effects, optimization may incorrectly attribute base-model errors to payload parameters. To address this, HALO adopts a two-stage identification strategy that first calibrates the base model and then estimates payload-related parameters under loaded conditions. This structured decomposition significantly improves identification stability and accuracy.

With the identified dynamics, we can train locomotion policies in simulation and deploy them zero-shot to hardware. In simulation, our approach demonstrates clear advantages over existing methods: against traditional SysID approaches such as CMA-ES, HALO yields more accurate estimates of physical parameters, especially under extreme load conditions. Compared to baseline strategies, our policies achieve lower motion-tracking errors. Real-world experiments further validate these results, showing that policies retain high precision and agility under heavy payloads across a range of challenging tasks. These experiments confirm that by reducing structured model bias before policy learning, our framework effectively narrows the sim-to-real gap while preserving agility and robustness. Specifically, under a heavy-loaded condition, HALO improves performance by $45.45\%$ to $73.33\%$ in the gait precision task and by $72.97\%$ in the in-place 90$^\circ$ yaw jumping task, respectively. Furthermore, HALO achieves 100\% success rate in challenging motion tracking tasks.

In summary, our contributions are summarized as follows:

\begin{itemize}
    \item \textbf{A Differentiable Two-Stage System Identification Framework:} We propose a trajectory-level gradient-based identification pipeline that decouples base-model calibration from payload parameter estimation, enabling stable and accurate parameter recovery.
    
    \item \textbf{Sensor-Minimal Data Collection:} By introducing a fixed-foot constraint, we collect trajectory data by reconstructing full-body states using only joint encoders, removing the dependency on torque sensing or motion capture systems.
    
    \item \textbf{Agile and Robust Dynamic Performance under Heavy Loads:} By reducing structured model bias prior to policy training, our approach enables zero-shot transfer of reinforcement learning policies to hardware. Extensive simulation and real-world experiments show more accurate parameter identification, improved motion tracking,  and superior agility.  
\end{itemize}

\section{RELATED WORK}

\subsection{Domain Randomization and Adaptation Strategy}
To bridge the sim-to-real gap without explicit model identification, researchers primarily focus on building policies that are either inherently robust to variations or capable of online compensation.

\paragraph{Domain Randomization (DR)}
DR remains the de facto standard for building robust policies by exposing the agent to a broad distribution of visual and physical variations during training \cite{related-work-DR1, related-work-DR2, related-work-DR4, related-work-DR5}. Recent advancements have evolved to adaptive schemes, such as curriculum-based or adversarial DR \cite{related-work-adversarial-dr, related-work-curriculum-dr}, which dynamically adjust parameter distributions. However, DR often requires heuristic tuning of randomization ranges. For heavy-payload tasks, excessive randomization typically leads to overly conservative behaviors, sacrificing the agility required to handle massive inertial shifts.

\paragraph{Adaptive Policy Learning}
Alternative approaches attempt to mitigate sim-to-real gap through real-time policy adjustment\cite{related-work-online-adaptive-policy2, related-work-online-adaptive-policy3, related-work-online-adaptive-policy1} or data-driven domain transfer\cite{related-work-offline-adaptive-policy1}. Some even incorporate lifelong learning mechanisms during deployment\cite{related-work-lifelong-adaptive-policy-1, related-work-lifelong-adaptive-policy-2}. However, the effectiveness of these methods is frequently bottlenecked by the requirement for high-quality supervisory signals, and they typically fail to provide the seamless zero-shot transition necessary for immediate field operation.

\paragraph{Data-driven Dynamics Learning}
Another class of method involves integrating learned residual model to bridge the gap between idealized simulation and real-world complexity\cite{related-work-dynamic-learning-1, intro-ASAP}. Instead of identifying physical parameters, these methods deploy neural ``patches'' to generate corrective efforts or actions for unmodeled dynamics. While these ``patch'' models perform well in the specific scenarios they were trained for, they often fail to generalize to diverse tasks.

\subsection{Physical Parameter Alignment via System Identification}
Rather than learning a robust or adaptive policy that covers a wide range of dynamics, system identification aims to minimize the sim-to-real gap by identifying the true physical parameters. 

\paragraph{Traditional and Sampling-based Methods} 
Traditional system identification relies on specialized hardware or offline calibration to estimate the physical parameters\cite{related-work-traditional-husky, related-work-traditional-football}. These approaches are often limited by noise sensitivity and the difficulty of measurement in real-world. More recently, sampling-based evolutionary algorithms have been employed to explore high-dimensional parameter spaces \cite{intro-ETH, related-work-sampling-based-2}. While these methods handle non-differentiable dynamics well, they typically suffer from low sample efficiency due to their reliance on derivative-free optimization, particularly in complex robotic systems.

\paragraph{Differentiable Simulation Methods}
Differentiable simulation \cite{related-work-mjx, related-work-brax} has emerged as a paradigm-shifting tool for parameter alignment by providing analytical gradients of system dynamics with respect to physical properties. Unlike derivative-free engines, differentiable simulators enable end-to-end gradient-based optimization, allowing for structured, sample-efficient updates in high-dimensional parameter spaces. This approach has demonstrated remarkable success in mitigating sim-to-real discrepancies across diverse robotic platforms, ranging from fixed-base manipulators \cite{related-work-dif-sim-fixed-base-manipulator1, related-work-dif-sim-fixed-base-manipulator2} to floating-base systems such as quadcopters \cite{related-work-dif-sim-quadcopter} and quadruped robots \cite{related-work-dif-sim-quadruped}. While recent breakthroughs have successfully optimized complex actuator parameters for a humanoid system \cite{kovalev2025achieving}, scaling this framework to heavy-payload humanoids—where massive inertial shifts and non-linear dynamics dominate—remains an open, high-stakes challenge that requires a more robust identification strategy.

\section{METHODS}

We consider a robot with known nominal model parameters that must execute locomotion skills while carrying an unknown payload with its upper limbs. Our goal is to first identify the effective model parameters under the payload and subsequently execute motion skills based on the identified dynamics.

Our method consists of three steps. 
(i) \textit{Data Collection}: We execute an exploration policy to control the robot under both unloaded and loaded conditions to collect state–action trajectory data. 
(ii) \textit{Gradient-Based Identification}: We estimate the effective model parameters via a two-stage gradient-based optimization procedure that minimizes the trajectory matching error under loaded dynamics. 
(iii) \textit{Heavy-Load Skill Acquisition}: Using the identified parameters, we train reinforcement learning policies for agile locomotion tasks, including balancing, dancing, and payload-carrying. Fig.~\eqref{fig:method_overview} provides an overview of the proposed method.

\subsection{Data Collection}

Accurate global state trajectories are typically difficult to obtain for a floating-base humanoid robot without external measurement systems such as motion capture (MoCap). To address this, we constrain one foot of the robot using 2 vises, enabling full-body trajectories to be reconstructed from joint-state measurements via forward kinematics. The same kinematic constraint is applied in the simulation to maintain consistency.
\begin{figure}[t] 
    \centering
    \includegraphics[width=0.7\linewidth]{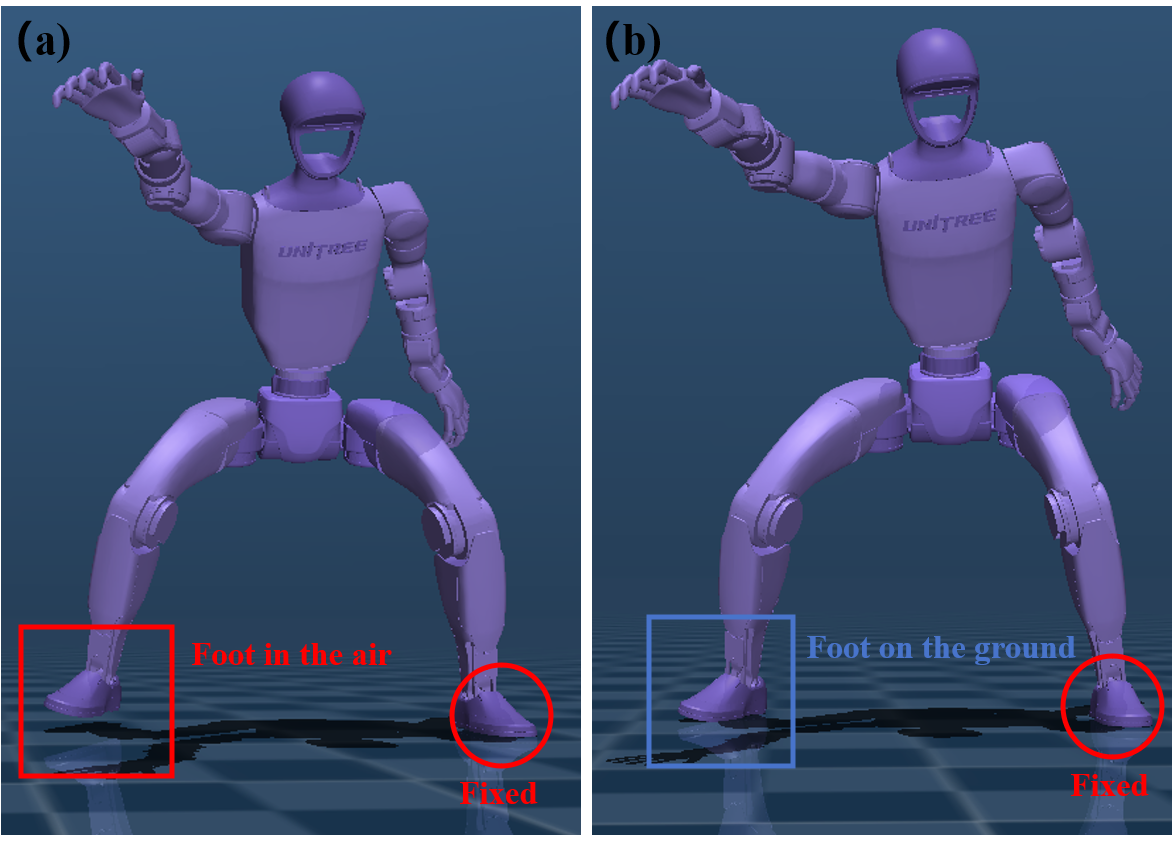}
    \caption{\textbf{Inconsistent left-right foot heights caused by sensor noise.}
    \textbf{(a)} In the original trajectory, the right foot position calculated via forward kinematics remains suspended in the air.
    \textbf{(b)} In the optimized trajectory, the right foot maintains physically consistent ground contact.}
    \label{fig:figure_method_1}
\end{figure}
We heuristically design a robot trajectory with the fixed foot and train an exploration policy to imitate it, following the procedure described in Sec.~\eqref{sec:method-rl}. Data is then collected on the physical robot using this policy, recording joint positions, and the corresponding control inputs. Here, the control input refers to the target joint positions commanded by the RL policy and tracked via a low-level PD controller. The policy is trained with nominal model parameters and a wide range of domain randomization to ensure stability under varying payload conditions. 
Additionally, due to sensor noise, directly recover body state from joint measurements may introduce accumulated drift, leading to inconsistent left–right foot heights, as illustrated in Figure~\eqref{fig:figure_method_1}. To correct this, at each timestep we solve a constrained quadratic program over the root and lower-body joint increments, minimally adjusting the configuration to eliminate foot-height discrepancies.

Because small changes in mass or CoM position can quickly cause the simulated trajectory to drift, only a small amount of data is needed to identify the main differences between the model and the real robot.

\subsection{Gradient-Based Identification}
\subsubsection{Problem Formulation}
Let $\boldsymbol{\theta} \in \mathbb{R}^d$ denote a set of physical parameters, including link mass, CoM position, inertia, and joint damping coefficients. 

The real-world system evolves according to unknown dynamics:
\begin{equation}
    s^{r}_{i+1} = \Phi_{\text{real}}(s^{r}_{i}, a^{r}_{i}; \boldsymbol{\theta}^{r}),
\end{equation}
where $s_i$, $a_i$, and $\boldsymbol{\theta}^{r}$ denote the system state, control input at timestep $i$, and the groundtruth parameters of the real system, respectively.

In simulation, we approximate the dynamics with a differentiable model:
\begin{equation}
    s^{s}_{i+1} = \Phi_{\text{sim}}(s^{s}_{i}, a^{r}_{i}; \boldsymbol{\theta}),
\end{equation}
where the control sequence is fixed to the recorded real-world actions $a^{r}_{i}$, and the initial state is synchronized:
$s^{s}_{0} = s^{r}_{0}.$

Given a collected real-world trajectory 
$\tau^{r} = \{s^{r}_0, a^{r}_0, \dots, s^{r}_N\}$,
our goal is to estimate parameters $\boldsymbol{\theta}$ such that the simulated trajectory 
$\tau^{s}(\boldsymbol{\theta}) = \{s^{s}_0, \dots, s^{s}_N\}$ 
matches the real trajectory.

We formulate system identification as a trajectory-matching optimization:

\begin{equation}
    \min_{\boldsymbol{\theta}} 
    \mathcal{L}_{\text{total}}(\boldsymbol{\theta})=\mathcal{L}_{\text{track}}(\boldsymbol{\theta})
    + \mathcal{R}(\boldsymbol{\theta}),
\end{equation}

\noindent where the tracking loss $\mathcal{L}_{\text{track}}(\boldsymbol{\theta})$ measures trajectory discrepancy between the simulated trajectories and the real-world reference trajectories in Cartesian space, and $\mathcal{R}(\boldsymbol{\theta})$ denotes regularization terms that enforce physical plausibility and prevent singular solutions.

The resulting optimization problem is solved via gradient-based methods by leveraging the differentiability of $\Phi_{\text{sim}}$. The parameters are updated iteratively using Gradient Descent (GD):

\begin{equation}
    \boldsymbol{\theta}_{k+1} = \boldsymbol{\theta}_k - \eta \nabla_{\boldsymbol{\theta}} \mathcal{L}_{\text{total}}(\boldsymbol{\theta}_k),
\end{equation}
where $k$ denotes the optimization iteration index, $\eta$ is the learning rate. 

\subsubsection{Tracking Loss}


For a complete real motion trajectory, we uniformly sample a batch of $B$ trajectory fragments. At each sampled starting point, we reset the robot's state in the simulation environment and execute a rollout of $N$ control steps to calculate loss. The tracking loss consists of two components: (1) the position loss for all bodies $\mathcal{B}$ in the robot, and (2) an additional loss for the upper-body subset $\mathcal{U} \subset \mathcal{B}$ weighted by $\alpha^u$. This additional term emphasizes accurate tracking of the body links most influenced by the payload. The tracking loss function is formulated as:


\begin{equation}
    \mathcal{L}_{\text{track}}(\boldsymbol{\theta}) = \mathcal{L}^{all}_{\text{track}}(\boldsymbol{\theta})+ \alpha^u\mathcal{L}^{upper}_{\text{track}}(\boldsymbol{\theta})
\end{equation}
\begin{equation}
    \mathcal{L}^{all}_{\text{track}}(\boldsymbol{\theta}) = \frac{1}{B} \sum_{b=1}^{B} \sum_{t=1}^{N}
        \sum_{j \in \mathcal{B}} \left\| \mathbf{x}_{b,t,j}^{s}(\boldsymbol{\theta}) - \mathbf{x}_{b,t,j}^{r} \right\|_2^2
\end{equation}
\begin{equation}
    \mathcal{L}^{upper}_{\text{track}}(\boldsymbol{\theta}) = \frac{1}{B} \sum_{b=1}^{B} \sum_{t=1}^{N}
        \sum_{j \in \mathcal{U}} \left\| \mathbf{x}_{b,t,j}^{s}(\boldsymbol{\theta}) - \mathbf{x}_{b,t,j}^{r} \right\|_2^2
\end{equation}

\noindent where:
\begin{itemize}
    \item $\mathcal{B}$ denotes the set of all robot bodies,
    \item $\mathcal{U} \subset \mathcal{B}$ denotes the subset of upper-body links (e.g., torso and hands),
    \item $\mathbf{x}_{b,t,j}^{s}(\boldsymbol{\theta}), \mathbf{x}_{b,t,j}^{r} \in \mathbb{R}^3$ denote the simulated and real Cartesian positions of body $j$ at timestep $t$ in batch $b$, which are components of the corresponding system states.
\end{itemize}


\subsubsection{Regularization Terms}

To prevent the identified parameters from deviating substantially from their nominal values, we introduce regularization terms that penalize large changes. 
Let $\mathcal{I}_{\text{link}}$ and $\mathcal{I}_{\text{joint}}$ denote the set of links and joints whose parameters are identified, respectively. 
The total regularization is then defined as a weighted sum of four components:

\begin{equation}
    \mathcal{R}_{\text{total}} 
        = \lambda_{\text{com}} \mathcal{R}_{\text{com}} 
        + \lambda_{\text{mass}} \mathcal{R}_{\text{mass}} 
        + \lambda_{\text{damp}} \mathcal{R}_{\text{damp}} 
        + \lambda_{\text{fric}} \mathcal{R}_{\text{fric}}.
\end{equation}

Each term is defined as follows:

{\small
\[
\begin{aligned}
    \mathcal{R}_{\text{com}}  &= \sum_{i \in \mathcal{I}_{\text{link}}} \| \mathbf{p}_i - \mathbf{p}_i^{\text{base}} \|_2^2 \\
    \mathcal{R}_{\text{mass}} &= \sum_{i \in \mathcal{I}_{\text{link}}} (m_i - m_i^{\text{base}})^2 \\
    \mathcal{R}_{\text{damp}} &= \sum_{k \in \mathcal{I}_{\text{joint}}} \phi(\alpha_{d,k}; 0.8, 1.2) \\
    \mathcal{R}_{\text{fric}} &= \sum_{k \in \mathcal{I}_{\text{joint}}} \phi(\alpha_{f,k}; 0.8, 1.2)
\end{aligned}
\]
}

Here, $\mathbf{p}_i$ denotes the CoM position of the $i$-th link relative to its parent link, and $m_i$ is the corresponding link mass. Their nominal values are given by $\mathbf{p}_i^{\text{base}}$ and $m_i^{\text{base}}$. 
The variables $\alpha_{d,k}$ and $\alpha_{f,k}$ represent the joint damping coefficient and static friction of the $k$-th joint, respectively. 

Accordingly, $\mathcal{R}_{\text{com}}$ and $\mathcal{R}_{\text{mass}}$ penalize deviations of link CoM positions and masses from their nominal values, while 
$\mathcal{R}_{\text{damp}}$ and $\mathcal{R}_{\text{fric}}$ softly constrain the joint damping and friction to remain within a reasonable interval using the box-constraint function:
\begin{equation}
    \phi(\alpha; l, u) = \max(0, l - \alpha)^2 + \max(0, \alpha - u)^2,
\end{equation}
which only penalizes values outside the interval $[l, u]$.


\subsubsection{Two-Stage System Identification}

Directly optimizing model parameters from nominal values using trajectories collected under loaded conditions can lead to biased estimates. 
In practice, humanoid robots exhibit Sim-to-Real discrepancies due to wear, calibration drift, and minor structural variations, which exist even in the absence of payloads. 
Without compensating for this nominal model mismatch, the optimization process may incorrectly attribute base-model errors to payload-related parameters.

We therefore adopt a two-stage identification procedure that separates base-model calibration from payload estimation.

\begin{enumerate}
    \item \textbf{Stage 1: Base Model Calibration.} 
    Using trajectories collected without payload, we optimize the full set of model parameters starting from their nominal values, 
    yielding a calibrated base model that reduces the inherent Sim-to-Real gap.

    \item \textbf{Stage 2: Payload Parameter Identification.} 
    With the calibrated model serving as initialization, we optimize only the payload-related parameters, 
    specifically the mass and its CoM positions of the links attached by the payload, i.e., torso and hands, 
    using trajectories collected under loaded conditions.
\end{enumerate}

In practice, restricting the second stage to mass and CoM parameters is sufficient to significantly reduce the Sim-to-Real gap in heavy-load scenarios, while avoiding the complexity of optimizing the inertia tensor. 



\subsection{Skill Acquisition Under Heavy-Load}
\label{sec:method-rl}
The identified parameters can be directly utilized for general simulation training. We adopt a motion imitation method based on mjlab~\cite{experiment-mjlab} to train RL policies for agile motion skills. 
A set of reference robot motion trajectories is pre-collected and used as imitation targets. 
In simulation, MLP-based policies are optimized using PPO to maximize a tracking reward that encourages alignment of joint states and body poses with the reference trajectories. After training, policies can be directly deployed on the real robot. 
Owing to the accurately identified model parameters, only minimal domain randomization is required during training, enabling zero-shot sim-to-real transfer of the learned skills to the physical robot.



\begin{table}[htbp]
  \centering
  \caption{Domain Randomization Settings.}
  \label{tab:domain_randomization}
  \footnotesize  
  \renewcommand{\arraystretch}{0.9}  
  \begin{tabular}{lcc}
    \toprule
    \textbf{Term} & \textbf{Nominal Range} & \textbf{Wide Range} \\
    \midrule
    Base CoM offset (m) & $\mathcal{U}(-0.03, 0.03)$ & $\mathcal{U}(-0.08, 0.08)$ \\
    Torso mass offset (kg) & $\mathcal{U}(-0.5, 0.5)$ & $\mathcal{U}(2.0, 8.0)$ \\
    Hands mass offset (kg) & $\mathcal{U}(-0.5, 0.5)$ & $\mathcal{U}(1.0, 4.0)$ \\
    Hands CoM offset (m) & $\mathcal{U}(-0.02, 0.02)$ & $\mathcal{U}(-0.05, 0.05)$ \\
    Encoder bias (rad) & $\mathcal{U}(-0.01, 0.01)$ & $\mathcal{U}(-0.01, 0.01)$ \\
    Foot friction & $\mathcal{U}(0.3, 1.2)$ & $\mathcal{U}(0.3, 1.2)$ \\
    \bottomrule
  \end{tabular}
\end{table}
\section{EXPERIMENT}
Our experiments aim to address the following key questions:
\begin{itemize}
    \item \textbf{Q1:} Can HALO outperform CMA-ES based methods for heavy-loaded humanoid SysID?
    \item \textbf{Q2:} Does HALO bring better performance compared to other methods for motion tracking tasks under heavy payloads?
    \item \textbf{Q3:} Does the two-stage method lead to higher real-world SysID accuracy?
    \item \textbf{Q4:} Can the parameters identified by HALO improve the robustness of sim-to-real transfer for heavy-loaded motion tracking tasks?
\end{itemize}
\subsection{Experiment overview}
The advantages of HALO are validated through both simulations (Secs. \eqref{sec:GD_Mujoco}, \eqref{sec:sim_motion_tracking}) and real-world experiments (Secs. \eqref{sec:real_sysid}, \eqref{sec:real experiment}). 

\textbf{Experiment Settings.} We use Unitree G1 to test performance of HALO, which stands approximately 132~cm tall and has a nominal mass of 35~kg. 
\begin{itemize}
    \item \textbf{Simulation Payload Setting}. For simulation, we design three different perturbation settings to manually perturb the physical parameters, simulating the sim2real gap introduced by payloads, as shown in Table~\eqref{tab:sim_id_results}.
    \item \textbf{Real-World Payload Setting}. We introduce payload in real-world by manually attaching 6 kg diving counterweights to the torso link, and 2 kg counterweights to each wrist link (hand), as shown in Figure~\eqref{fig:teaser}(a). 
\end{itemize}

\subsection{HALO vs. CMA-ES: Performance under Extreme Payload Perturbations}
\label{sec:GD_Mujoco}
To answer \textbf{Q1}, we collect data in simulation using the perturbed values, while setting the nominal values as the optimization start. The regularization terms were not enabled in simulation experiment. HALO utilizes learning rates of $0.03$ for mass and $0.0002$ for CoM position. CMA-ES is configured with a population size of 10. Both methods undergo 2,000 iterations, reaching complete convergence. For CMA-ES, we take experiments using five different random seeds, meaning different initial exploration paths. The perturbation settings and results are illustrated at Table~\eqref{tab:sim_id_results}.

\begin{table*}[htbp]
    \centering
    \caption{Convergence comparison between CMA-ES and HALO. Results are reported as mean $\pm$ standard deviation. Groundtruth values (e.g., $+6.000$, $+2.4000$) denote the incremental offsets added to the absolute nominal values. Bold values indicate performance closer to the ground truth. While CMA-ES converges under light payloads, it exhibits sensitivity to random seeds under heavy payloads.}
    \label{tab:sim_id_results}
    \scriptsize
    \newcolumntype{C}{>{\centering\arraybackslash}X}
    \begin{tabularx}{\textwidth}{l C C C C C C} 
        \toprule
        \textbf{Method} & 
        \makecell[c]{\textbf{Torso}\\ \textbf{Mass(kg)}} & 
        \makecell[c]{\textbf{Left Hand}\\ \textbf{Mass(kg)}} & 
        \makecell[c]{\textbf{Right Hand}\\ \textbf{Mass (kg)}} & 
        \makecell[c]{\textbf{Torso CoM}\\ \textbf{pos x (m)}} & 
        \makecell[c]{\textbf{Torso CoM}\\ \textbf{pos y (m)}} & 
        \makecell[c]{\textbf{Torso CoM}\\ \textbf{pos z (m)}} \\ 
        \midrule
        
        \rowcolor{gray!20} \multicolumn{7}{l}{\textit{Absolute Nominal Values}} \\ 
        & $7.8180$ & $0.2546$ & $0.2546$ & $0.0020$ & $0.0003$ & $0.1845$ \\ 
        \midrule
        
        \rowcolor{gray!20} \multicolumn{7}{l}{\textit{Perturbation Setting 1}} \\ 
        Groundtruth & $+6.0000$ & $+2.4000$ & $+2.0000$ & $+0.0100$ & $+0.0200$ &$+0.0500$ \\
        CMA-ES & $+6.0366 \pm {\scriptstyle 0.0128}$ & $+2.4009 \pm {\scriptstyle 0.0012}$ & $+1.9970 \pm {\scriptstyle 0.0012}$ & $\mathbf{+0.0099 \pm {\scriptstyle 0.0000}}$ & $+0.0199 \pm {\scriptstyle 0.0001}$ & $+0.0505 \pm {\scriptstyle 0.0004}$  \\
        \textbf{HALO} & $\mathbf{+5.9989 \pm {\scriptstyle 0.0000}}$ & $\mathbf{+2.4005 \pm {\scriptstyle 0.0000}}$ & $\mathbf{+1.9973 \pm {\scriptstyle 0.0000}}$ & $+0.0098 \pm {\scriptstyle 0.0000}$ & $\mathbf{+0.0199 \pm {\scriptstyle 0.0000}}$ & $\mathbf{+0.0505 \pm {\scriptstyle 0.0000}}$ \\
        \midrule
        
        \rowcolor{gray!20} \multicolumn{7}{l}{\textit{Perturbation Setting 2}} \\ 
        Groundtruth & $+9.0000$ & $+3.2000$ & $+2.8000$ & $+0.0200$ & $+0.0400$ &$+0.0700$ \\
        CMA-ES & $\mathbf{+8.9988 \pm {\scriptstyle 0.0321}}$ & $+3.2055 \pm {\scriptstyle 0.0093}$ & $+2.7935 \pm {\scriptstyle 0.0184}$ & $\mathbf{+0.0200 \pm {\scriptstyle 0.0000}}$ & $\mathbf{+0.0400 \pm {\scriptstyle 0.0000}}$ & $+0.0667 \pm {\scriptstyle 0.0002}$ \\
        \textbf{HALO} & $+9.0918 \pm {\scriptstyle 0.0000}$ & $\mathbf{+3.1981 \pm {\scriptstyle 0.0000}}$ & $\mathbf{+2.7982 \pm {\scriptstyle 0.0000}}$ & $+0.0199 \pm {\scriptstyle 0.0000}$ & $+0.0492 \pm {\scriptstyle 0.0000}$ & $\mathbf{+0.0668 \pm {\scriptstyle 0.0000}}$ \\
        \midrule

        \rowcolor{gray!20} \multicolumn{7}{l}{\textit{Perturbation Setting 3}} \\ 
        Groundtruth & $+12.0000$ & $+3.5000$ & $+3.0000$ & $+0.0200$ & $+0.0400$ & $+0.0700$\\
        CMA-ES & $+12.4289 \pm {\scriptstyle 0.9244}$ & $+4.7885 \pm {\scriptstyle 2.7441}$ & $+3.6660 \pm {\scriptstyle 1.3323}$ & $-0.0616 \pm {\scriptstyle 0.1824}$ & $-0.0248 \pm {\scriptstyle 0.1449}$ & $+1.2141 \pm {\scriptstyle 2.5582}$  \\
        \textbf{HALO} & $\mathbf{+11.9871 \pm {\scriptstyle 0.0000}}$ & $\mathbf{+3.5014 \pm {\scriptstyle 0.0000}}$ & $\mathbf{+3.0016 \pm {\scriptstyle 0.0000}}$ & $\mathbf{+0.0200 \pm {\scriptstyle 0.0000}}$ & $\mathbf{+0.0400 \pm {\scriptstyle 0.0000}}$ & $\mathbf{+0.0701 \pm {\scriptstyle 0.0000}}$  \\
        \bottomrule
    \end{tabularx}
\end{table*}
 Experimental results demonstrate that under light payload perturbations, both HALO and CMA-ES methods achieve satisfactory convergence performance with comparable accuracy(e.g., Perturbation Setting 1,2). However, under extremely heavy-loaded conditions (e.g., Perturbation Setting 3), CMA-ES using certain seeds converge to local optima with physically unreasonable values, whereas our gradient-based method robustly converges to the vicinity of the reference parameters. These findings validate the superior optimization stability of our HALO for humanoid SysID under payload conditions.

\subsection{Impact of Accurate Payload Identification}
\label{sec:sim_motion_tracking}
To answer \textbf{Q2}, we evaluate the proposed framework under payload dynamics, specifically employing the configuration defined as \textbf{Perturbation Setting 1} in Table~\eqref{tab:sim_id_results}. We conducted motion tracking experiments in mjlab using sequences from AMASS\cite{experiment-datasets-AMASS}, LAFAN1\cite{experiment-datasets-lafan} and HuB\cite{experiment-datasets-hub}. The evaluation suite comprises 20 distinct sequences categorized into two difficulty levels: (i) \textit{Steady-state tasks} (e.g., walking with waving hands) and (ii) \textit{High-agility tasks} (e.g., side kicking), with 10 motions per category. 

Adopting the benchmarking setting established in \cite{related-work-sampling-based-2}, we compare our method against two representative base-lines: \textbf{(1) Wide Range DR (WDR),} as illustrated in Table~\eqref{tab:domain_randomization}, (col. 2). WDR uses the nominal model. \textbf{(2) Calibrated Mass (CM).} CM use the calibrated mass to simulating manually weighing the load, but the CoM positions are still nominal values. The DR range applied to CM and HALO is specified in Table~\eqref{tab:domain_randomization}, (col. 1). Finally, all resulting policies are validated on the ground-truth perturbed model to benchmark their tracking precision.

Following the evaluation framework established in \cite{han2025kungfubot2}, performance is quantified by three key metrics: \textit{Global Mean Per-Joint Position Error} ($E_{\text{g-mpjpe}}$, mm) for global alignment, \textit{Mean Per-Joint Position Error} ($E_{\text{mpjpe}}$, mm) for local precision, \textit{Velocity Error} ($E_{\text{vel}}$, m/s) for global root velocities error.

As summarized in Table \eqref{tab:results}, HALO consistently outperforms both baselines across all metrics. This superiority stems from its precise identification of payload mass and CoM offsets. In contrast, WDR relies on a broad randomization range which leads to overly conservative tracking, while CM fails to account for the dynamic variations induced by CoM displacement. By explicitly identifying the physical parameters of these rigid bodies, HALO maintains superior tracking precision compared to existing baselines.

\begin{table}[ht]
    \centering
    \caption{Experimental results of motion tracking performance across various tasks.}
    \label{tab:results}
    \footnotesize 
    \begin{tabular}{lccc} 
        \toprule
        \textbf{Method} & $E_{\text{g-mpjpe}}$ & $E_{\text{mpjpe}}$ & $E_{\text{vel}}$ \\
        \midrule
        \rowcolor{gray!20}
        \multicolumn{4}{l}{\textit{Steady-state tasks}} \\
        WDR & $94.91 \pm {\scriptstyle 3.21} $ & $70.54 \pm {\scriptstyle 2.13}$ & $0.157 \pm {\scriptstyle 0.008}$ \\
        CM & $85.87 \pm {\scriptstyle 2.54} $ & $56.33 \pm {\scriptstyle 2.27}$ & $0.130 \pm {\scriptstyle 0.006}$ \\
        \textbf{HALO} & $\mathbf{52.47} \pm {\scriptstyle 2.11}$ & $\mathbf{44.87}\pm {\scriptstyle 1.82}$ & $\mathbf{0.098}\pm {\scriptstyle 0.008}$ \\
        \midrule
        \rowcolor{gray!20}
        \multicolumn{4}{l}{\textit{High-agility tasks}} \\
        WDR & $132.89 \pm {\scriptstyle 2.89}$ & $81.54 \pm {\scriptstyle 2.43}$ & $0.136 \pm {\scriptstyle 0.016}$ \\
        CM & $92.82 \pm {\scriptstyle 3.21} $ & $67.06 \pm {\scriptstyle 3.21}$ & $0.125 \pm {\scriptstyle 0.011}$ \\
        \textbf{HALO} & $\mathbf{78.83} \pm {\scriptstyle 2.55}$ & $\mathbf{59.54} \pm {\scriptstyle 2.07}$ & $\mathbf{0.106} \pm {\scriptstyle 0.009}$ \\
        \bottomrule
    \end{tabular}
\end{table}
\subsection{Effectiveness of the Two-Stage Strategy}
\label{sec:real_sysid}
In this section, we compare two-stage method with one-stage baseline. The one-stage method directly identifies masses and CoM positions of loaded torso link and hand links, using only the loaded trajectories collected from the real-world. We collect three trajectories with payload and three trajectories without payload, and take the average of the final converged values as identification results. Reference masses are reported as the sum of the nominal masses and the masses of additional counterweights for loaded links. We conduct an initial assessment by evaluating how close the identified masses are to reference masses, because the groundtruth values of CoM positions cannot be obtained in realworld.

\begin{figure}[t]
    \centering
    \includegraphics[width=\linewidth]{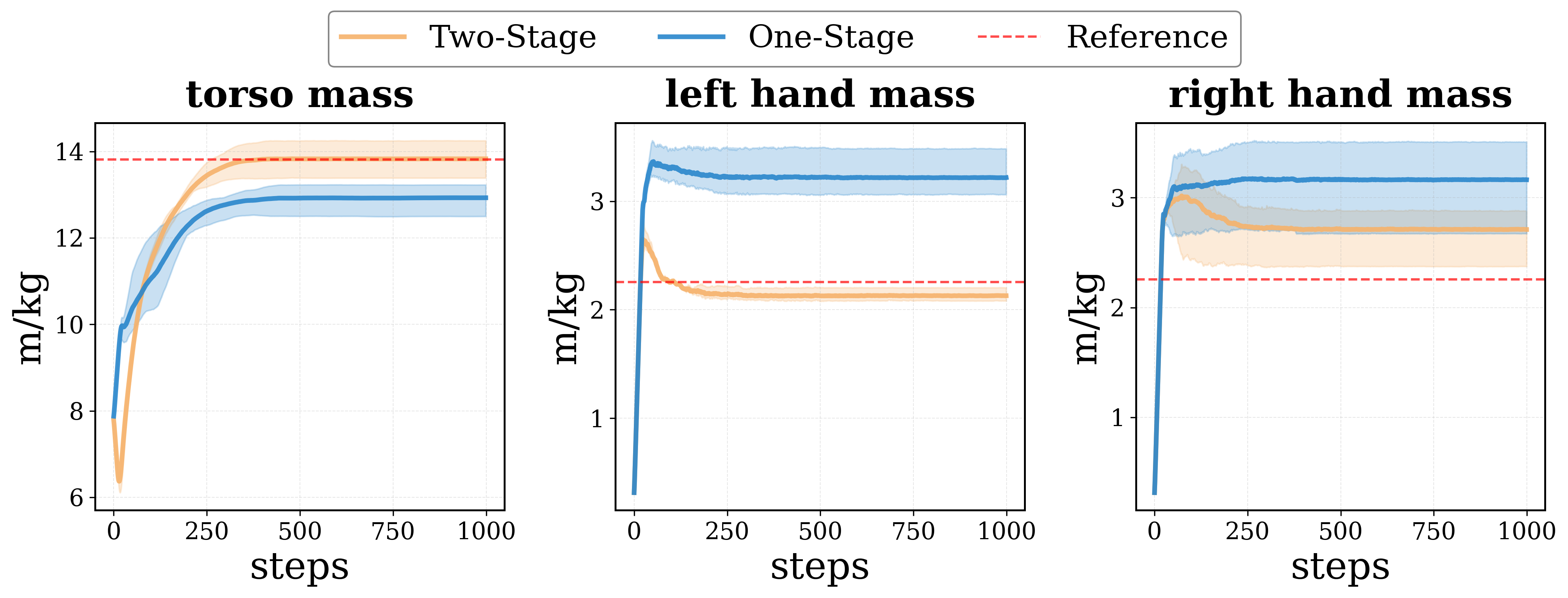}
    \caption{\textbf{Comparison of convergence performance between two-stage and one-stage methods.} The proposed Two-Stage method HALO (orange) demonstrates superior accuracy, converging significantly closer to the estimated reference values (red dashed lines) compared to the One-Stage baseline (blue). }
    \label{fig:real_convergence}
\end{figure}

The comparison of convergence performance between two-stage method (showing only the second stage) and one-stage method is illustrated in  Fig.~\eqref{fig:real_convergence}. The final results are summarized in Table~\eqref{tab:params real}. Compared to the one-stage method, the two-stage method demonstrates superior convergence accuracy, yielding mass estimates that are significantly closer to the reference values. This improvement suggests that our coarse-to-fine strategy effectively decouples global modeling errors from local payload variations. Furthermore, we will prove the effectiveness of the identified parameters by real-world downstream tasks as shown in Sec.~\eqref{sec:real experiment}.


\begin{table}[htbp]
\centering
\caption{Summary of identified parameters.}
\label{tab:params real}
\resizebox{\columnwidth}{!}{ 
\begin{tabular}{lccc} 
\toprule
\textbf{Method} & \textbf{Torso Mass (kg)} & \textbf{L-Hand Mass (kg)} & \textbf{R-Hand Mass (kg)} \\
\midrule
\rowcolor{gray!20}
Reference & 13.82 & 2.25 & 2.25 \\
One-Stage & $12.93 \pm 0.32$ & $3.22 \pm 0.23$ & $3.16 \pm 0.44$ \\
\textbf{HALO} & $\mathbf{13.83}\pm 0.43$ & $\mathbf{2.13}\pm 0.07$ & $\mathbf{2.71}\pm 0.29$ \\
\bottomrule
\end{tabular}
} 
\end{table}

\begin{figure}[t]
    \centering
    \includegraphics[width=\linewidth]{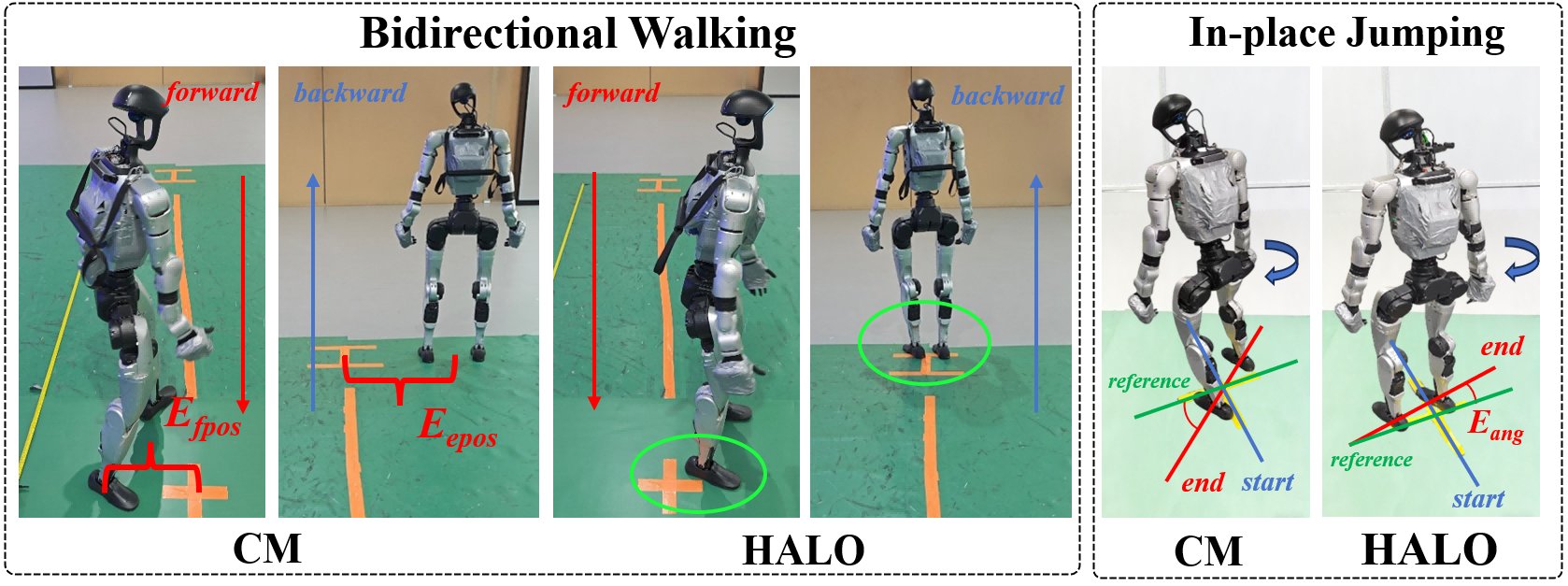}
    \caption{ \textbf{Experiment settings and results of scenario 1.} The metrics we defined are shown in the image, HALO outperforms the baselines across all metrics.}
    \label{fig:real walk}
\end{figure}

\subsection{Real-World Validation Under Heavy Payload Conditions}
\label{sec:real experiment}
Based on the optimized robot model identified via the HALO framework, we evaluate the effectiveness of our approach across different real-world tasks. Consistent with Sec.~\eqref{sec:sim_motion_tracking}, we compare against two baseline methods, WDR and CM, on two scenarios:

\begin{itemize}
    \item \textbf{Scenario 1: Quantitative Analysis of Fundamental Locomotion.}     
    \item \textbf{Scenario 2: Evaluation of High-Agility Motion Tasks.} 
\end{itemize}

In Scenario 1, to evaluate sim-to-real consistency under heavy payload conditions, we benchmark our method on two locomotion tasks: bidirectional walking (forward-and-backward) and in-place 90° yaw jumping. 
To formally quantify the sim-to-real gap, we introduce three evaluation metrics: \textit{Maximum Forward Position Error} ($E_{fpos}$, m), defined as the absolute discrepancy in forward displacement during the walking task; \textit{End-point Residual Error} ($E_{epos}$, m), representing the positional discrepancy at the final stopping state of the walking task; and \textit{Angular Tracking Error} ($E_{ang}$, deg) for assessing orientation deviation during jumping maneuvers. Notably, the WDR baseline failed to initiate the takeoff phase during the jumping task, primarily due to the inability to manage the heavy payload dynamics; its angular deviation was thus recorded as a task failure.

The experimental results are visualized in Fig.~\eqref{fig:real walk}, with the corresponding numerical analysis summarized in Table~\eqref{tab:real world walk}. Compared to the baseline methods (WDR and CM), HALO significantly mitigates sim-to-real positional drift. Specifically, our approach reduces $E_{fpos}$ by 73.33\% and 45.45\%, and decreases $E_{epos}$ by 70.79\% and 42.22\%, respectively. Furthermore, in the jumping task, HALO diminishes the angular tracking error $E_{ang}$ by 72.97\% compared to the CM baseline. These empirical results validate the efficacy of the HALO framework in enhancing the robustness and deployment fidelity of heavy-loaded humanoid systems in real-world environments.


While all methods achieve comparable performance in simulation, WDR fails to generate sufficient vertical impulse in the real-world. CM partially completes the maneuver but exhibits significant orientation deviation. In contrast, HALO achieves the lowest angular error, demonstrating that accurate identification of inertial and actuation parameters is critical for transferring explosive whole-body maneuvers.

\begin{table}[htbp]
\centering
\caption{Performance of Walking Forward and Backward.}
\label{tab:real world walk}
\begin{tabular}{lcccc} 
\toprule
\textbf{Method} & \textbf{$E_{fpos}$} & \textbf{$E_{epos}$} & \textbf{$E_{ang}$}\\
\midrule

WDR        & $0.45 \pm {\scriptstyle 0.13} $ & $0.89 \pm {\scriptstyle 0.15} $ & $N/A$  \\
CM             & $0.22 \pm {\scriptstyle 0.09} $ & $0.45 \pm {\scriptstyle 0.13} $ & $41.8 \pm {\scriptstyle 3.6} $  \\
\textbf{HALO}        & $\mathbf{0.12} \pm {\scriptstyle 0.03}$ & $\mathbf{0.26} \pm {\scriptstyle 0.07}$ & $\mathbf{11.3} \pm {\scriptstyle 2.1} $  \\


\bottomrule
\end{tabular}
\end{table}

In Scenario 2, to assess the performance on high-agility tasks, we benchmark our method on three challenging maneuvers: \textbf{swallow balancing, side kicking, and roundhouse kicking}~\cite{experiment-datasets-hub}. These tasks are characterized by complex whole-body coordination and substantial dynamic challenges to the robot's equilibrium. Success is quantified by the deviation from the nominal trajectory; specifically, a trial is deemed successful if the robot executes the maneuver without significant divergence from the reference motion. Detailed experimental behaviors are provided in the supplementary videos. Our results (Table \eqref{tab:success_rate}) indicate that the proposed method consistently outperforms baselines in terms of reliability.  


\begin{table}[htbp] 
\centering 
\caption{Comparison of success rates for performing challenging maneuvers between HALO and baselines.} 
\label{tab:success_rate}
\begin{tabularx}{\linewidth}{lXXX} 
\toprule
\small
\textbf{Method} & \textbf{Swallow Balancing} & \textbf{Side Kicking} & \textbf{Roundhouse Kicking}
\\
\midrule
WDR        & $0/10$ & $0/10$ & $0/10$ \\
CM      & $5/10$ & $7/10$ & $5/10$ \\
\textbf{HALO}   & $\mathbf{10/10}$ & $\mathbf{10/10}$ & $\mathbf{10/10}$ \\
\bottomrule
\end{tabularx}
\end{table}


\section{CONCLUSIONS}
In this paper, we present HALO, a framework with two-stage system identification based on differentiable simulation to address the significant sim-to-real gap for heavy-loaded humanoid robots. By leveraging analytical gradients from MuJoCo XLA, our approach achieves accurate and efficient estimation of robot physical parameters and payload mass distribution with only joint encoders, eliminating the need for expensive torque sensors or motion capture systems. The proposed two-stage optimization strategy achieves improved SysID accuracy, and enables robust sim-to-real transfer for heavy-loaded humanoid motion skills.







%



\bibliographystyle{plain} 
\bibliography{references}

\end{document}